\documentclass[10pt,twocolumn,letterpaper]{article}

\usepackage{cvpr}
\usepackage{times}
\usepackage{epsfig}
\usepackage{graphicx}
\usepackage{amsmath}
\usepackage{amssymb}
\usepackage{kotex}

\usepackage[numbers,sort]{natbib}
\usepackage{xcolor, colortbl}
\usepackage{tabularx}
\usepackage{multirow}
\usepackage{enumitem}

\newcolumntype{C}[1]{>{\centering\let\newline\\\arraybackslash\hspace{0pt}}p{#1}}

\def\ie{\emph{i.e.}}
\def\eg{\emph{e.g.}}
\def\etal{\emph{et al.}}
\def\wrt{\emph{w.r.t.}}

\definecolor{brown}{rgb}{0.65, 0.16, 0.16}
\definecolor{purp}{rgb}{0.65, 0.16, 0.65}

\newcommand{\jiwo}[1]{{\color{purp}{[#1]}}}

\newcommand{\Fig}[1]{Fig.~\ref{fig:#1}}
\newcommand{\Sec}[1]{Sec.~\ref{sec:#1}}
\newcommand{\Eq}[1]{Eq.~(\ref{eq:#1})}
\newcommand{\Tbl}[1]{Table~\ref{tab:#1}}

\usepackage[pagebackref=true,breaklinks=true,letterpaper=true,colorlinks,bookmarks=false]{hyperref}

\cvprfinalcopy 

\def\cvprPaperID{2973} 

\begin{document}

\title{Weakly Supervised Learning of Instance Segmentation with Inter-pixel Relations}

\author{Jiwoon Ahn\\
DGIST, \ Kakao Corp.\\
{\tt\small corey.ahn@kakaocorp.com}
\and
Sunghyun Cho\thanks{Co-corresponding authors.}\\
DGIST\\
{\tt\small scho@dgist.ac.kr}
\and
Suha Kwak\footnotemark[1]\\
POSTECH\\
{\tt\small suha.kwak@postech.ac.kr}
}

\maketitle


\begin{abstract}

This paper presents a novel approach for learning instance segmentation with image-level class labels as supervision.
Our approach generates pseudo instance segmentation labels of training images, which are used to train a fully supervised model.
For generating the pseudo labels, we first identify confident seed areas of object classes from attention maps of an image classification model, and propagate them to discover the entire instance areas with accurate boundaries. 
To this end, we propose IRNet, which estimates rough areas of individual instances and detects boundaries between different object classes.
It thus enables to assign instance labels to the seeds and to propagate them within the boundaries so that the entire areas of instances can be estimated accurately.
Furthermore, IRNet is trained with inter-pixel relations on the attention maps, thus no extra supervision is required.
Our method with IRNet achieves an outstanding performance on the PASCAL VOC 2012 dataset, surpassing not only previous state-of-the-art trained with the same level of supervision, but also some of previous models relying on stronger supervision.

\end{abstract}


\begin{figure*} [!t]
\centering
\includegraphics[width = 0.95 \textwidth]{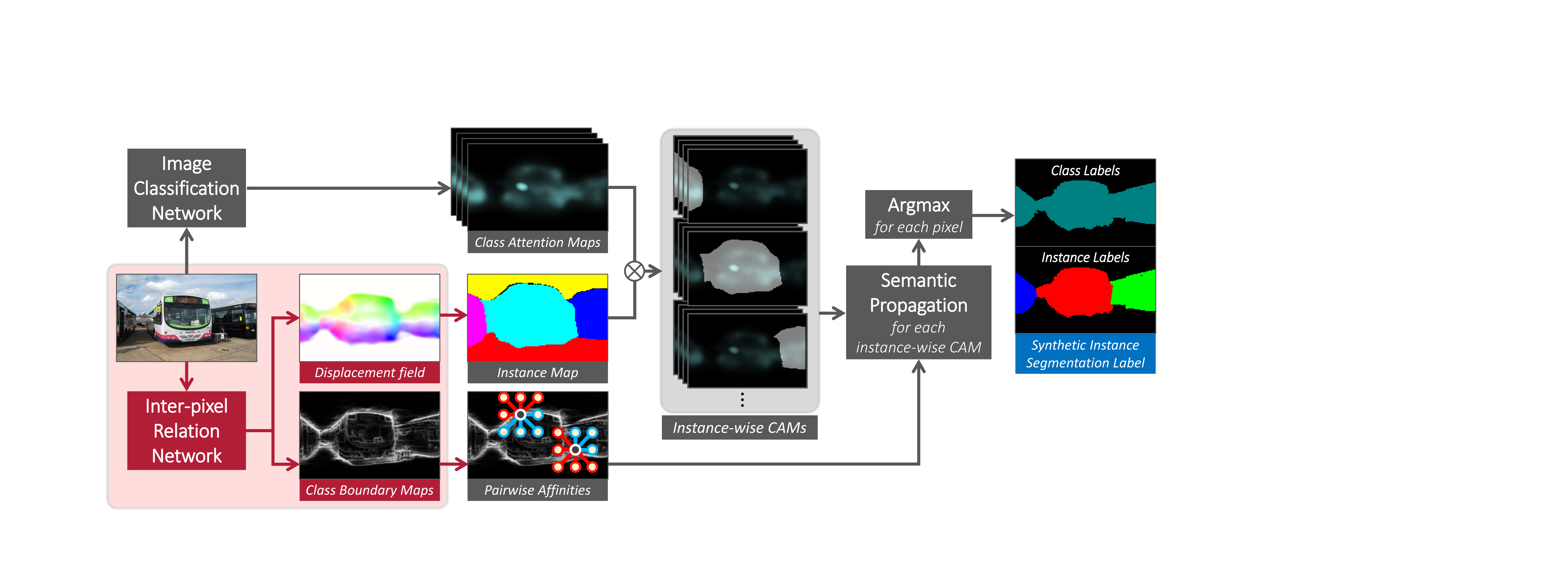}
\caption{
Overview of our framework for generating pseudo instance segmentation labels.
} 
\vspace{-0.3cm}
\label{fig:label_synthesis}
\end{figure*}

\section{Introduction}
\label{sec:intro}
Instance segmentation is a task that jointly estimates class labels and segmentation masks of individual objects.
As in other visual recognition tasks, supervised learning of Convolutional Neural Networks (CNNs) has driven recent advances in instance segmentation~\cite{Sds,DAICVPR15,Dai16_SDS,mask_rcnn,instancecut,Semiconv,Dai2016,Liu2016}.
Due to the data-hungry nature of deep CNNs, this approach demands an enormous number of training images with groundtruth labels, which are given by hand in general. 
However, manual annotation of instance-wise segmentation masks is prohibitively time-consuming, which results in existing datasets limited in terms of both class diversity and the amount of annotated data.
It is thus not straightforward to learn instance segmentation models that can handle diverse object classes in the real world.

One way to alleviate this issue is weakly supervised learning that adopts weaker and less expensive labels than instance-wise segmentation masks as supervision.
Thanks to low annotation costs of weak labels, approaches in this category can utilize more training images of diverse objects, although they have to compensate for missing information in weak labels.
For instance segmentation, bounding boxes have been widely used as weak labels since they provide every property of objects except shape~\cite{SDI,Cutnpaste}.
However, it is still costly to obtain box labels for a variety of classes in a large number of images as they are manually annotated.

To further reduce the annotation cost, one may utilize image-level class labels for learning instance segmentation since such labels are readily available in large-scale image classification datasets, \eg, ImageNet~\cite{Russakovsky2015}.
Furthermore, although image-level class labels indicate only the existence of object classes, they can be used to derive strong cues for instance segmentation, called \emph{Class Attention Maps} (CAMs)~\cite{Cam,Oquab15,Wei_2017_CVPR,Selvaraju_2017_ICCV}.
A CAM roughly estimates areas of each class by investigating the contribution of local image regions to the classification score of the class.
However, CAMs cannot be directly utilized as supervision for instance segmentation since they have limited resolution, often highlight only partial areas of objects, and most importantly, cannot distinguish different instances of the same class.
To resolve this issue, a recent approach~\cite{PRM} incorporates CAMs with an off-the-shelf segmentation proposal technique~\cite{MCG}, which however has to be trained separately on an external dataset with additional supervision.
In this paper, we present a novel approach for learning instance segmentation using image-level class labels, which outperforms the previous state-of-the-art trained with the same level of supervision~\cite{PRM} and even some of approaches relying on stronger supervision~\cite{SDI,Sds}.
Moreover, it requires neither additional supervision nor any segmentation proposals unlike the previous approaches~\cite{Sds,PRM}.
Our method generates pseudo instance segmentation labels of training images given their image-level labels and trains a known CNN model with the pseudo labels.
For generating the pseudo labels, it utilizes CAMs, but as mentioned earlier, they can neither distinguish different instances nor find entire instance areas with accurate boundaries.

To overcome these limitations of CAMs, we introduce \emph{Inter-pixel Relation Network} (IRNet) that is used to estimate two types of additional information complementary to CAMs: a class-agnostic instance map and pairwise semantic affinities.
A class-agnostic instance map is a rough instance segmentation mask without class labels nor accurate boundaries.
On the other hand, the semantic affinity between a pair of pixels is a confidence score for class equivalence between them.
By incorporating instance-agnostic CAMs with a class-agnostic instance map, we obtain instance-wise CAMs, which are in turn enhanced by propagating their attention scores to relevant areas based on the semantic affinities between neighboring pixels.
After the enhancement, a pseudo instance segmentation label is generated by selecting the instance label with the highest attention score in the instance-wise CAMS at each pixel.
The entire procedure for label synthesis is illustrated in \Fig{label_synthesis}.

IRNet has two branches estimating an instance map and semantic affinities, respectively. 
The first branch predicts a displacement vector field where a 2D vector at each pixel indicates the centroid of the instance the pixel belongs to.
The displacement field is converted to an instance map by assigning the same instance label to pixels whose displacement vectors point at the same location.
The second branch detects boundaries between different object classes.
Pairwise semantic affinities are then computed from the detected boundaries in such a way that two pixels separated by a strong boundary are considered as a pair with a low semantic affinity.
Furthermore, we found that IRNet can be trained effectively with inter-pixel relations derived from CAMs.
Specifically, we collect pixels with high attention scores and train IRNet with the displacements and class equivalence between the collected pixels.
Thus, no supervision in addition to image-level class labels is required.

The contribution of this paper is three-fold:
\vspace{-2mm}
\begin{itemize}[leftmargin=5mm] 
\itemsep=-1mm
\item We propose a new approach to identify and localize instances with image-level supervision through class-agnostic instance maps. 
This enables instance segmentation without off-the-shelf segmentation proposals.
\item We propose a new way to learn and predict semantic affinities between pixels with image-level supervision through class boundary detection, which is more effective and efficient than previous work~\cite{affinitynet}.
\item On the PASCAL VOC 2012 dataset~\cite{Pascalvoc}, our model substantially outperforms the previous state-of-the-art trained with the same level of supervision~\cite{PRM}. 
Also, it even surpasses previous models based on stronger supervision like SDI~\cite{SDI} that uses bounding box labels and SDS~\cite{Sds}, an early model that uses full supervision. 
\end{itemize}

\section{Related Work}
\label{sec:relatedwork}

This section reviews semantic and instance segmentation models closely related to our method.
We first introduce weakly supervised approaches for the two tasks, and discuss models that are based on ideas similar with the displacement field and pairwise semantic affinity of our framework.

\noindent \textbf{Weakly Supervised Semantic Segmentation:}
For weak supervision of semantic segmentation,
various types of weak labels such as bounding boxes~\cite{Boxsup, Wssl}, scribbles~\cite{scribblesup, Vernaza2017}, and points~\cite{Bearman16} have been utilized. 
In particular, image-level class labels have been widely used as weak labels since they require minimal or no effort for annotation~\cite{oh17cvpr,Huang_2018_CVPR,Tokmakov16,Hong2017_webly,Wsl,wildcat,PRM,Wei_2017_CVPR,affinitynet}. 
Most approaches using the image-level supervision are based on CAMs~\cite{Cam,Oquab15,Selvaraju_2017_ICCV} that roughly localize object areas by drawing attentions on discriminative parts of object classes.
However, CAMs often fail to reveal the entire object areas with accurate boundaries.
To address this issue, extra data or supervision have been exploited to obtain additional evidences like saliency~\cite{oh17cvpr,Huang_2018_CVPR}, motion in videos~\cite{Tokmakov16, Hong2017_webly} and class-agnostic object proposals~\cite{Wsl}.
Recent approaches tackle the issue without external information 
by mining complementary attentions iteratively~\cite{Wei_2017_CVPR,Huang_2018_CVPR} or propagating CAMs based on semantic affinities between pixels~\cite{affinitynet}.

\noindent \textbf{Weakly Supervised Instance Segmentation:}
For instance segmentation, bounding boxes have been widely used as weak labels.
Since a bounding box informs the exact location and scale of an object, weakly supervised models using box labels focus mainly on estimating object shapes.
For example, in~\cite{SDI}, GraphCut is incorporated with generic boundary detection~\cite{HED} to better estimate object shapes by considering boundaries.
Also, in~\cite{Cutnpaste}, an object shape estimator is trained by adversarial learning~\cite{GAN} so that a pseudo image generated by cutting and pasting the estimated object area to a random background looks realistic.
Meanwhile, weakly supervised instance segmentation with image-level class labels has been rarely studied since this is a significantly ill-posed problem where supervision does not provide any instance-specific information.
To tackle this challenging problem, a recent approach~\cite{PRM} detects peaks of class attentions to identify individual instances and combines them with high-quality segmentation proposals~\cite{MCG} to reveal entire instance areas.
However, the performance of the method heavily depends on that of the segmentation proposals, which have to be trained with extra data with high-level supervision.
In contrast, our approach requires neither off-the-shelf proposals nor additional supervision and it surpasses the previous work~\cite{PRM} by a substantial margin.


\noindent \textbf{Pixel-wise Prediction of Instance Location:}
Pixel-wise prediction of instance location has been proven to be effective for instance segmentation in literature. 
In~\cite{Liang_sds} the coordinates of the instance bounding box each pixel belongs to are predicted in a pixel-wise manner so that pixels with similar box coordinates are clustered as a single instance mask. 
This idea is further explored in~\cite{kendall2017multi, Semiconv}, which predict instance centroids instead of box coordinates. 
Our approach based on the displacement field share the same idea with~\cite{kendall2017multi, Semiconv}, but it requires only image-level supervision while the previous approaches are trained with instance-wise segmentation labels.


\noindent \textbf{Semantic Affinities Between Pixels:}
Pairwise semantic affinities between pixels have been used to enhance the quality of semantic segmentation.
In~\cite{randomwalk_net, ChengCVPR2017}, CNNs for semantic segmentation are incorporated with a differentiable module computing a semantic affinity matrix of pixels, and trained in an end-to-end manner with full supervision.
In \cite{randomwalk_net}, a predicted affinity matrix is used as a transition probability matrix for random walk, while in \cite{ChengCVPR2017}, it is embedded into a convolutional decoder~\cite{deconvnet} to encourage local pixels to have the same labels during inference.
Recently, a weakly supervised model has been proposed to learn pairwise semantic affinities with image-level class labels~\cite{affinitynet}.
This model predicts a high-dimensional embedding vector for each pixel, and the affinity between a pair of pixels is defined as the similarity between their embedding vectors.
Our approach shares the same motivation with~\cite{affinitynet}, but our IRNet can learn and predict affinities more effectively and efficiently by detecting class boundaries. 

\section{Class Attention Maps}
\label{sec:cams}

CAMs play two essential roles in our framework.
First, they are used to define seed areas of instances, which are propagated later to recover the entire instance areas as in~\cite{sec,affinitynet}.
Second, they are a source of supervision for learning IRNet;
by exploiting CAMs carefully, we extract reliable inter-pixel relations, from which IRNet is trained.
To generate CAMs for training images, we adopt the method of~\cite{Cam} using an image classification CNN with global average pooling followed by a classification layer.
Given an image, the CAM of a groundtruth class $c$ is computed by
\begin{equation}
    M_c (\mathbf{x}) = \frac{\phi_c^\top  f (\mathbf{x})}{\max_{\mathbf{x}} \phi_c^\top  f (\mathbf{x})},
\end{equation}
where $f$ is a feature map from the last convolution layer of the CNN, $\mathbf{x}$ is a 2D coordinate on $f$, and $\phi_c$ is the classification weights of the class $c$.
Also, CAMs for irrelevant classes are fixed to a zero matrix.
We adopt ResNet50~\cite{resnet} as the classification network, 
and reduce the stride of its last downsampling layer from 2 to 1 to prevent CAMs from further resolution drop.
As a result, the width and height of CAMs are $1/16$ of those of the input image.

\section{Inter-pixel Relation Network}
\label{sec:irnet}
\begin{figure} [!t]
\begin{center}
\includegraphics[width=1.0 \linewidth] {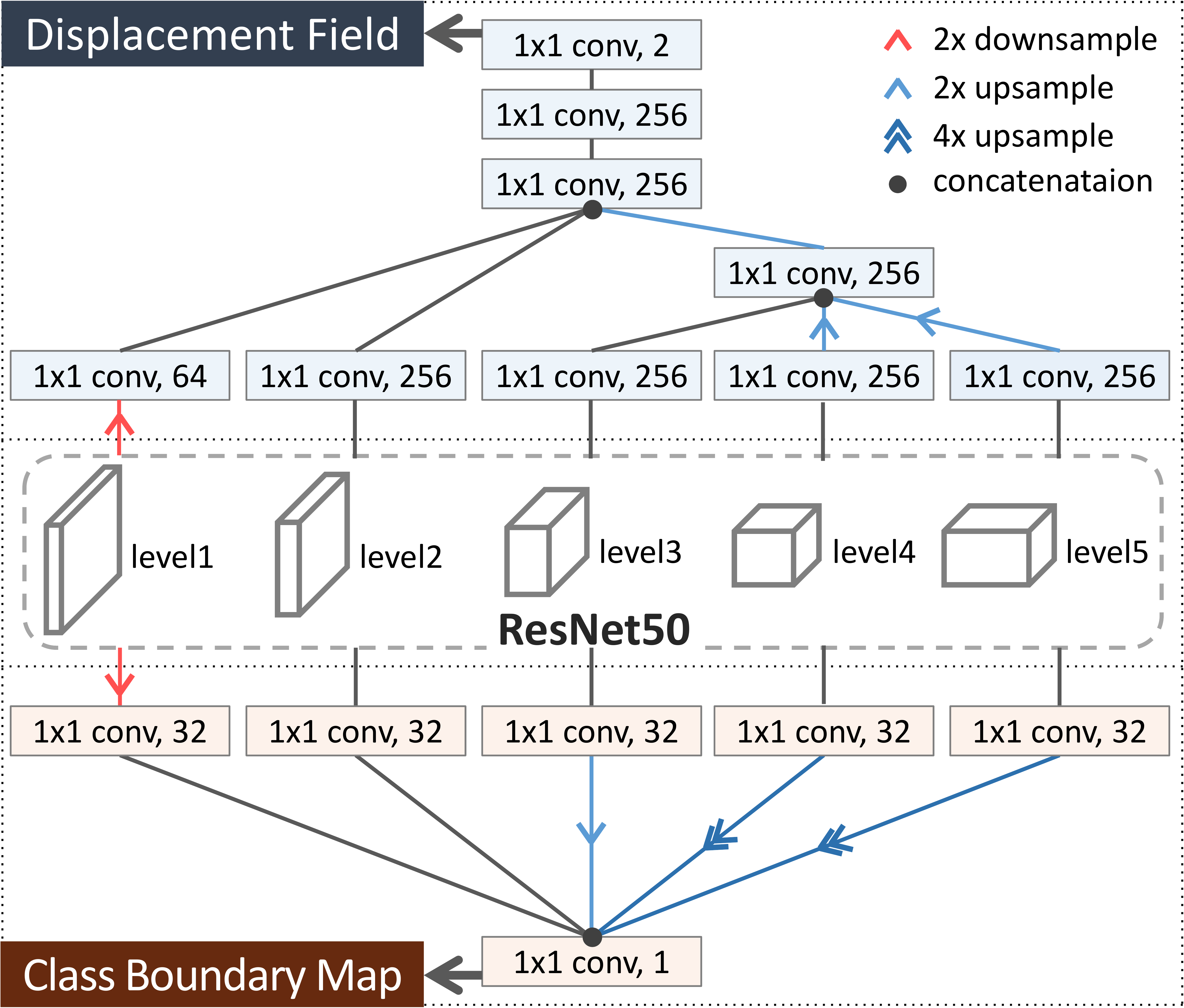}
\end{center}
\vspace{-0.3cm}
\caption{
Overall architecture of IRNet.
} 
\label{fig:net}
\vspace{-0.1cm}
\end{figure}

IRNet aims to provide two types of information: a displacement vector field and a class boundary map, both of which are in turn used to estimate pseudo instance masks from CAMs.
This section describes the IRNet architecture 
and the strategy for learning the model using CAMs as supervision.
How to use IRNet for pseudo label generation will be illustrated in \Sec{labelsynth}.

\subsection{IRNet Architecture}
IRNet has two output branches that predict a displacement vector field and a class boundary map, respectively.
Its architecture is illustrated in \Fig{net}.
The two branches share the same ResNet50 backbone, which is identical to that of the classification network in \Sec{cams}.
As inputs, both branches take feature maps from all the five levels\footnote{A level means a group of residual units sharing the same output size in~\cite{resnet}. However, in our backbone, the output sizes of \textsf{\scriptsize level4} and \textsf{\scriptsize level5} are identical since the stride of the last downsampling layer is reduced to 1.} of the backbone.
All the convolution layers of both branches are followed by group normalization~\cite{GroupNorm} and ReLU except the last layer.
Details of both branches are described below.

\noindent \textbf{Displacement Field Prediction Branch:}
A 1$\times$1 convolution layer is first applied to each input feature map, and the number of channels is reduced to 256 if it is larger than that.
On top of them, we append a top-down path way~\cite{lin2017_fpn} to merge all the feature maps iteratively in such a way that low resolution feature maps are upsampled twice, concatenated with those of the same resolution, and processed by a 1$\times$1 convolution layer.
Finally, from the last concatenated feature map, a displacement field is decoded through three 1$\times$1 convolution layers, whose output has two channels.

\noindent \textbf{Boundary Detection Branch:}
We first apply 1$\times$1 convolution to each input feature map for dimensionality reduction.
Then the results are resized, concatenated, and fed into the last 1$\times$1 convolution layer, which produces a class boundary map from the concatenated features. 

\subsection{Inter-pixel Relation Mining from CAMs}
\label{sec:interpixelrel}

Inter-pixel relations are the only supervision for training IRNet, thus it is important to collect them reliably.
We define two kinds of relations between a pair of pixels: the displacement between their coordinates and their class equivalence.
The displacement can be easily computed by a simple subtraction, but the class equivalence is not since pixel-wise class labels are not given in our weakly supervised setting.

\begin{figure} [!t]
\centering
\includegraphics[width = 0.9 \linewidth]{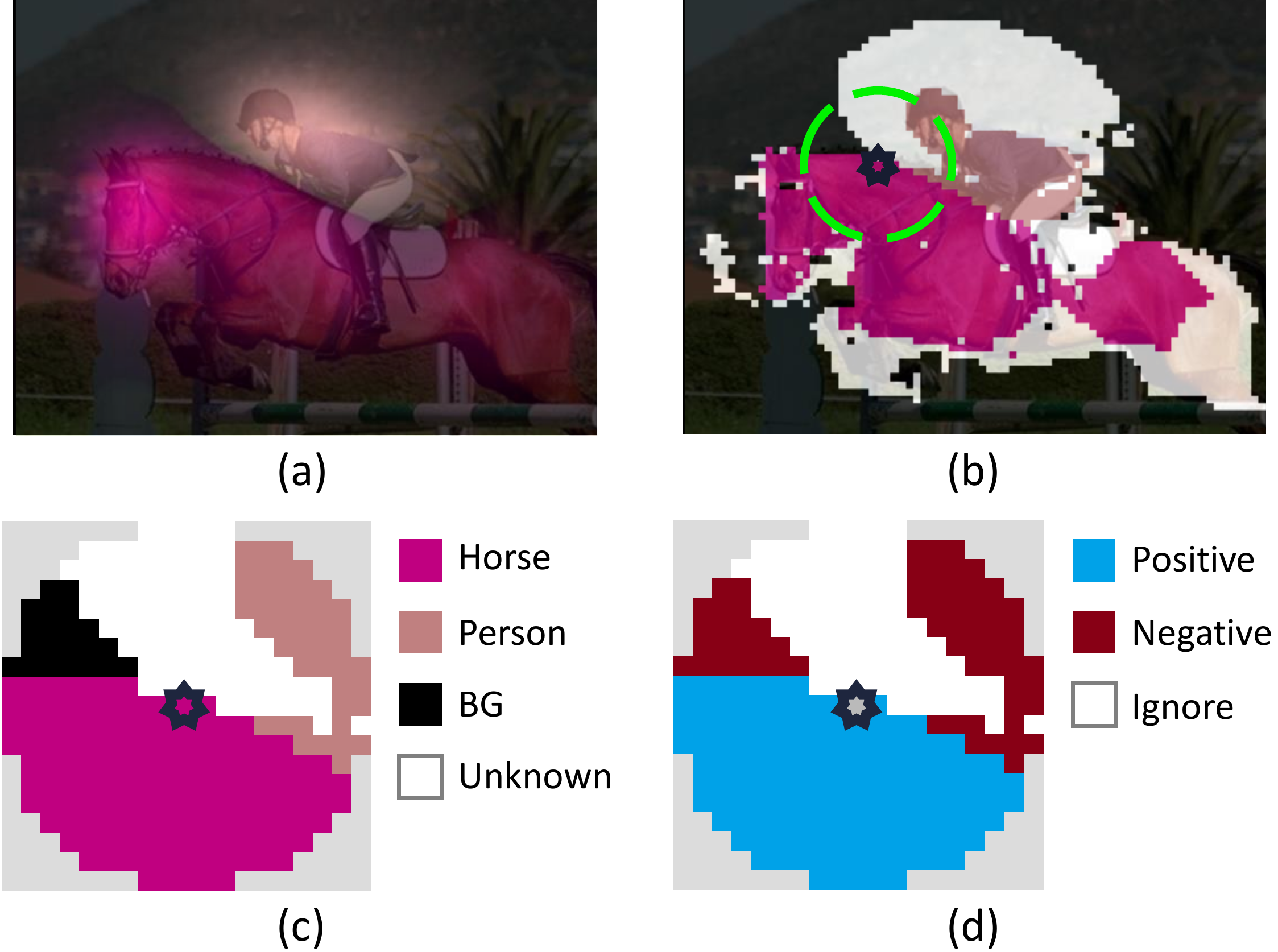}
\caption{
Visualization of our inter-pixel relation mining process. (a) CAMs. (b) Confident areas of object classes. (c) Pseudo class label map within a local neighborhood. (d) Class equivalence relations between the center and the others.
}
\vspace{-0.1cm}
\label{fig:pair}
\end{figure}
Thus, we carefully exploit CAMs to predict pixel-wise pseudo class labels and obtain reliable class equivalence relations from them.
The overall procedure of our method is illustrated in \Fig{pair}.
Since CAMs are blurry and often inaccurate, we first identify areas with confident foreground/background attention scores.
Specifically, we collect pixels with attention scores larger than $0.3$ as foreground pixels, and smaller than $0.05$ as background pixels.
Note that we do not care pixels outside of confident areas during the process.
Each confident area is then refined by dense CRF~\cite{Fullycrf} to better estimate object shapes.
After that, we construct a pseudo class map $\hat{M}$ by choosing the class with the best score for each pixel.
Finally, we sample pairs of neighboring pixels from the refined confident areas, and categorize them into two sets $\mathcal{P}^+$ and $\mathcal{P}^-$ according to their class equivalence by
\begin{align}
    \mathcal{P} &= \big\{(i, j) \mid \lVert \mathbf{x}_i - \mathbf{x}_j \rVert_2 < \gamma, \forall i\neq j \big\},
    \label{eq:pair_set}
    \\
    \mathcal{P}^+ &=  \big\{(i, j) \mid \hat{M}(\mathbf{x}_i) = \hat{M}(\mathbf{x}_j), (i, j) \in \mathcal{P} \big\},
    \label{eq:pair_set_plus}
    \\
    \mathcal{P}^- &= \big\{(i, j) \mid \hat{M}(\mathbf{x}_i) \neq \hat{M}(\mathbf{x}_j), (i, j) \in \mathcal{P} \big\},
    \label{eq:pair_set_minus}
\end{align}
where $\gamma$ is a radius limiting the maximum distance of a pair. 
We further divide $\mathcal{P}^+$ into $\mathcal{P}^+_\text{fg}$ and $\mathcal{P}^+_\text{bg}$, a set of foreground pairs and that of background pairs, respectively.

\subsection{Loss for Displacement Field Prediction}

The first branch of IRNet predicts a displacement vector field $\mathcal{D}\in\mathbb{R}^{w\times h\times 2}$, where each 2D vector points at the centroid of the associated instance.
Although ground truth centroids are not given in our setting, 
we argue that $\mathcal{D}$ can be learned implicitly with displacements between pixels of the same class.
There are two conditions for $\mathcal{D}$ to be a displacement field.
First, for a pair of pixel locations $\mathbf{x}_i$ and $\mathbf{x}_j$ belonging to the same instance, their estimated centroids must be identical, \ie, 
$\mathbf{x}_i + \mathcal{D}(\mathbf{x}_i) = \mathbf{x}_j + \mathcal{D}(\mathbf{x}_j)$.
Second, by the definition of centroid, $\sum_\mathbf{x} \mathcal{D(\mathbf{x})} = 0$ for each instance.

To satisfy the first condition, we first assume that a pair of nearby pixels $(i, j)\in \mathcal{P}^+$ is likely to be of the same instance since they are sampled within a small radius $\gamma$.
Then, given such a pair $(i, j)$, our goal is to approximate their image coordinate displacement $\hat{\delta}(i, j) = \mathbf{x}_j - \mathbf{x}_i$ with their difference in $\mathcal{D}$ denoted by $\delta(i, j) = \mathcal{D}(\mathbf{x}_i) - \mathcal{D}(\mathbf{x}_j)$.
In the ideal case where $\delta = \hat\delta$, it will hold that $\mathbf{x}_i + \mathcal{D}(\mathbf{x}_i) = \mathbf{x}_j + \mathcal{D}(\mathbf{x}_j)$ for all $(i, j)$ of the same instance.
This implies that $\mathcal{D}(\mathbf{x})$ is the displacement vector indicating the corresponding centroid.
For learning $\mathcal{D}$ with the inter-pixel relations obtained in \Sec{interpixelrel}, 
we minimize $L_1$ loss between $\delta(i, j)$ and $\hat\delta(i, j)$:
\begin{equation}
    \mathcal{L}^\mathcal{D}_\text{fg} = \frac{1}{| \mathcal{P}^+_\text{fg} |} \sum_{(i, j)\in \mathcal{P}^+_\text{fg}} \Big| \delta(i, j) - \hat\delta(i, j) \Big|.
    \label{eq:displacement_loss}
\end{equation}

The second condition, on the other hand, is not explicitly encouraged by \Eq{displacement_loss}.
However, we argue that IRNet can still learn to predict displacement vectors pointing to rough centroids of instances due to the randomness of initial network parameters. Intuitively speaking, initial random displacement vectors are already likely to satisfy the second condition, and the training of IRNet converges to a local minimum that still satisfies the condition. A similar phenomenon is observed in~\cite{Semiconv}.
Displacement vectors are then further refined by subtracting the mean of $\mathcal{D}$ from $\mathcal{D}$.

Also, we eliminate trivial centroid estimation from background pixels since the centroid of background is indefinite and may interfere with the above process.
For the purpose, we minimize the following loss for background pixels:
\begin{equation}
    \mathcal{L}^\mathcal{D}_\text{bg} = \frac{1}{| \mathcal{P}^+_\text{bg} |} \sum_{(i, j)\in \mathcal{P}^+_\text{bg}} | \delta(i, j) |.
\end{equation}

\subsection{Loss for Class Boundary Detection}
Given an image, the second branch of IRNet detects boundaries between different classes, and the output is denoted by $\mathcal{B}\in [0, 1]^{w\times h}$.
Although no ground truth labels for class boundaries are given in our setting, we can train the second branch with class equivalence relations between pixels through a Multiple Instance Learning (MIL) objective.
The key assumption is that a class boundary exists somewhere between a pair of pixels with different pseudo class labels.

To implement this idea, we express the semantic affinity between two pixels in terms of the existence of a class boundary.
For a pair of pixels $\mathbf x_i$ and $\mathbf x_j$,
we define their semantic affinity $a_{i j}$ as: 
\begin{equation}
    a_{i j} = 1 - \max_{k \in \Pi_{i j}}{\mathcal{B} (\mathbf{x}_k)}
\label{eq:affinity_def}
\end{equation}
where $\Pi_{i j}$ is a set of pixels on the line between $\mathbf x_i$ and $\mathbf x_j$.
We utilize class equivalence relations between pixels as supervision for learning $a_{i j}$.
Specifically, the class equivalence between two pixels is represented as a binary label whose value is 1 if their pseudo class labels are the same and 0 otherwise.
The affinity is then learned by minimizing cross-entropy between the one-hot vector of the binary affinity label and the predicted affinity in \Eq{affinity_def}:
\begin{align}
    \mathcal{L}^\mathcal{B} = & - \sum_{(i, j)\in \mathcal{P}^+_\text{fg}} \frac{\log a_{i j}}{2| \mathcal{P}^+_\text{fg} |} \ \ - \sum_{(i, j)\in \mathcal{P}^+_\text{bg}}  \frac{\log  a_{i j}}{2 | \mathcal{P}^+_\text{bg} |} \nonumber \\
    & - \sum_{(i, j)\in \mathcal{P}^-} \frac{\log( 1 - a_{i j} ) } {| \mathcal{P}^- |} \label{eq:boundary_objective}
\end{align}
where three separate losses are aggregated after normalization since populations of $\mathcal{P}^+_\text{fg}$, $\mathcal{P}^+_\text{bg}$, and $\mathcal{P}^-$ are significantly imbalanced in general.
Through the loss in \Eq{boundary_objective}, we can learn $\mathcal{B}$ implicitly with inter-pixel class equivalence relations.
In this aspect, \Eq{boundary_objective} can be regarded as a MIL objective where $\Pi_{i j}$ is a bag of potential boundary pixels.

\subsection{Joint Learning of the Two Branches}
The two branches of IRNet are jointly trained by minimizing all the losses we defined previously at the same time:
\begin{equation}
    \mathcal{L} = \mathcal{L}^\mathcal{D}_\text{fg} + \mathcal{L}^\mathcal{D}_\text{bg} + \mathcal{L}^\mathcal{B}. \label{eq:loss_final}
\end{equation}
Note that the above loss is class-agnostic since $\mathcal{P}^+$ and $\mathcal{P}^-$ only consider class equivalence between pixels rather than their individual class labels. 
This allows our approach to utilize more inter-pixel relations per class and helps to improve the generalization ability of IRNet.


\section{Label Synthesis Using IRNet}
\label{sec:labelsynth}

\begin{figure} [!t]
\centering
\includegraphics[width = 0.97 \linewidth, trim={0 1.3cm 0 0}, clip]{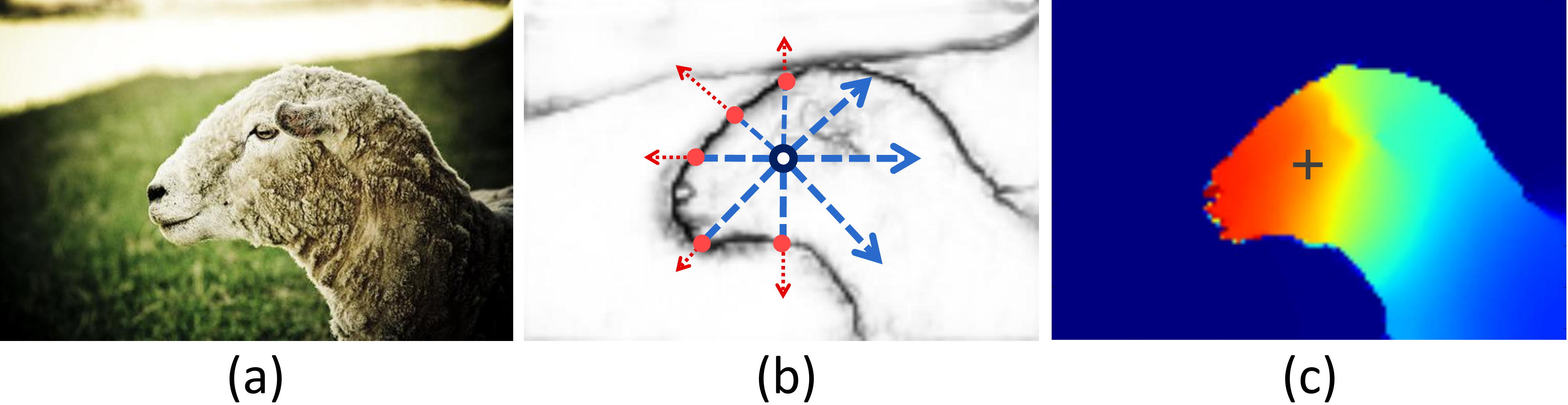}
\caption{Deriving pairwise semantic affinities from a class boundary map. (left) Input Image. (center) A class boundary map. (right) Label propagation from the center after random walks.
} 
\label{fig:aff}
\end{figure}

To synthesize pseudo instance labels, the two outputs $\mathcal{D}$ and $\mathcal{B}$ of IRNet are converted to a class-agnostic instance map and pairwise affinities, respectively.
Among them, semantic affinities can be directly derived from $\mathcal{B}$ by \Eq{affinity_def} as illustrated in \Fig{aff}, while the conversion of $\mathcal{D}$ is not straightforward due to its inaccurate estimation.
This section first describes how $\mathcal{D}$ is converted to an instance map, then how to generate pseudo instance segmentation labels with the instance map and semantic affinities.

\subsection{Generating Class-agnostic Instance Map}
\label{sec:class_agnostic_instance_map}

\begin{figure} [!t]
\centering
\includegraphics[width = 0.97 \linewidth, trim={0 1.4cm 0 0},clip]{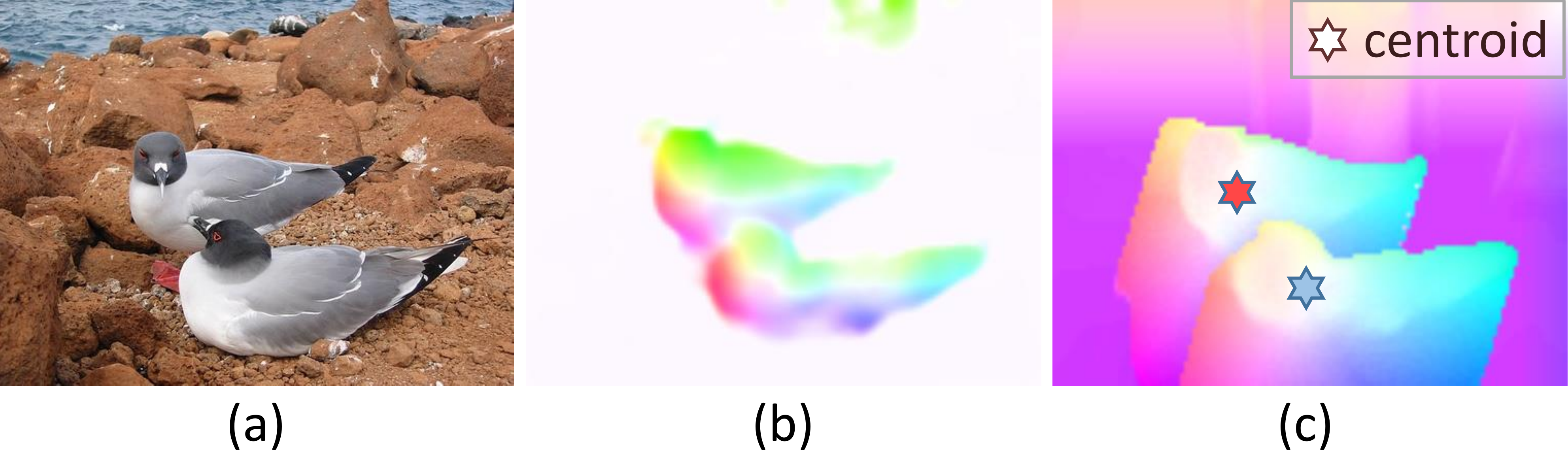}
\caption{Detecting instance centroids. (left) Input image. (center) An initial displacement field. (right) A refined displacement field and detected centroids.
}
\label{fig:centroid}
\end{figure}

A class-agnostic instance map $I$ is a $w\times h$ 2D map, each element of which is the instance label associated with the element.
If $\mathcal{D}$ is estimated with perfect accuracy, $I$ can be obtained simply by grouping pixels whose displacement vectors point at the same centroid. 
However, $\mathcal{D}$ often fails to predict the exact offsets to centroids since IRNet is trained with incomplete supervision derived from CAMs.
To address this issue, $\mathcal{D}$ is refined iteratively by 
\begin{eqnarray}
\mathcal{D}_{u+1}(\mathbf{x}) = \mathcal{D}_u(\mathbf{x}) + \mathcal{D}\left(\mathbf{x} + \mathcal{D}_u(\mathbf{x})\right) \ \ \forall \mathbf{x},\label{eq:displacement_update}
\end{eqnarray}
where $u$ is an iteration index and $\mathcal{D}_0$ is the initial displacement field given by IRNet. 
Each displacement vector is refined iteratively by adding the displacement vector at the currently estimated centroid location.
As displacement vectors near centroids tend to be almost zero in magnitude, the refinement converges within a finite number of iterations.
The effect of the refinement is demonstrated in \Fig{centroid}.

Since centroids estimated via the refined $\mathcal{D}$ are still scattered in general,
we consider a small group of neighboring pixels, instead of a single coordinate, as a centroid.
To this end, we first identify pixels whose displacement vectors in $\mathcal{D}$ have small magnitudes, and regard them as candidate centroids since pixels around a true centroid will have near zero displacement vectors.
Then each connected component of the candidates is considered as a centroid. 
Note that the candidates tend to be well grouped into a few connected components since displacement vectors change smoothly within a local neighborhood as can be seen in \Fig{centroid}.

\subsection{Synthesizing Instance Segmentation Labels}
\begin{figure} [!t]
\centering
\includegraphics[width = 1 \linewidth]{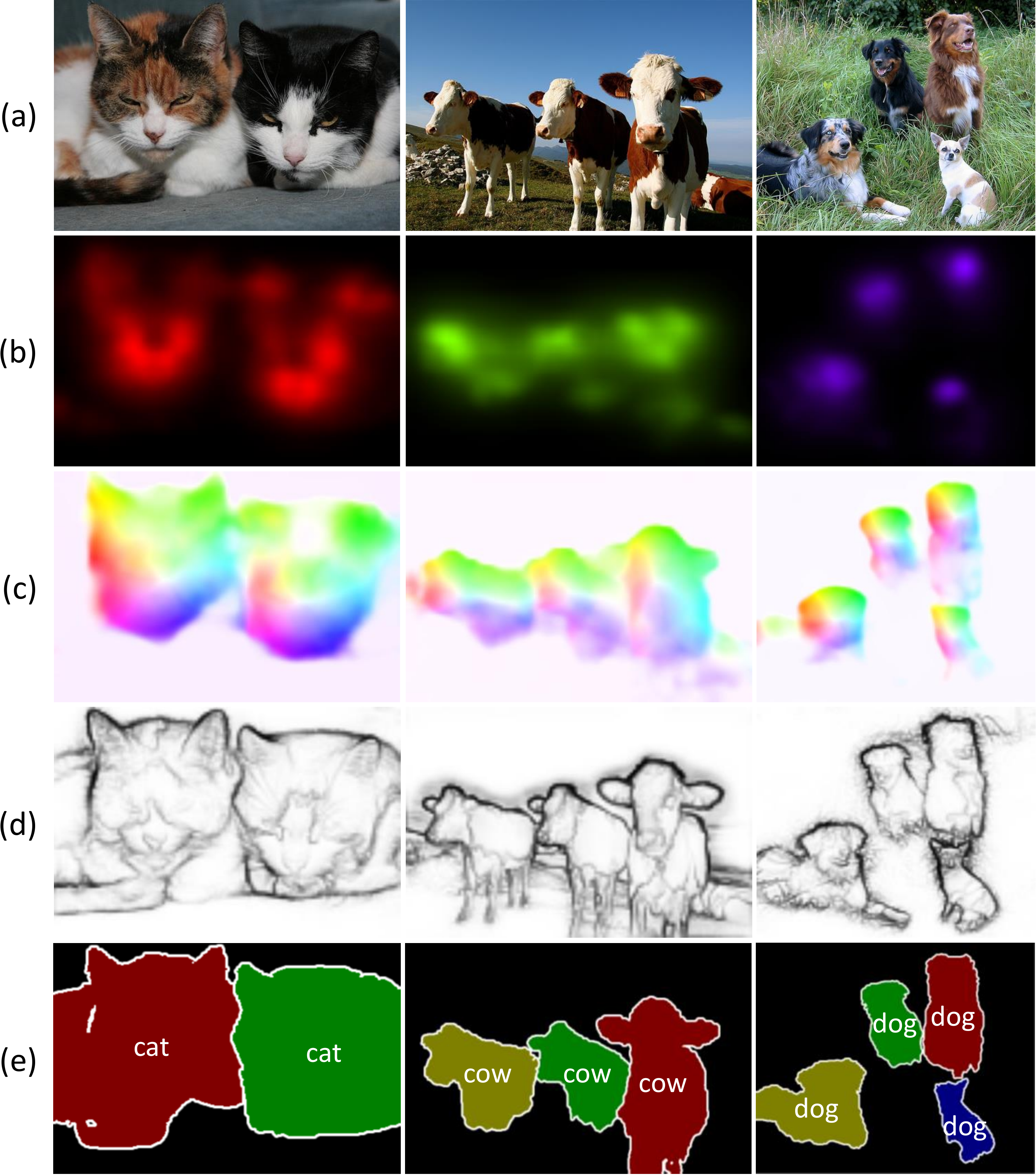}
\caption{Examples of pseudo instance segmentation labels on the PASCAL VOC 2012 \emph{train} set. (a) Input image. (b) CAMs. (c) Displacement field. (d) Class boundary map. (e) Pseudo labels.
} 
\label{fig:qualitative_labels}
\end{figure}
For generating pseudo instance masks, we first combine CAMs with a class-agnostic instance map as follows:
\begin{equation}
    \bar{M}_{c k}( \mathbf{x}) = 
    \begin{cases}
        M_c (\mathbf x) & \text{if } I(\mathbf x) = k, \\
        0              & \text{otherwise,}
    \end{cases}
\end{equation}
where $\bar{M}_{c k}$ is the instance-wise CAMs of class $c$ and instance $k$.
Each instance-wise CAM is refined individually by propagating its attention scores to relevant areas.
Specifically, the propagation is done by random walk, whose transition probability matrix is derived from the semantic affinity matrix $A=[a_{ij}]\in \mathbb{R}^{wh \times wh}$ as follows:
\begin{eqnarray}
    T = S^{-1} A^{\circ \beta}, \ \ \textrm{where} \ \ S_{i i} = \sum_j a_{i j}^\beta \label{eq:trans_mat}
\end{eqnarray}
and $A^{\circ \beta}$ is $A$ to the Hadamard power of $\beta$ and $S$ is a diagonal matrix for row-normalization of $A^{\circ \beta}$.
Also, $\beta>1$ is a hyper-parameter for smoothing out affinity values in $A$.
The random walk propagation with $T$ is then conducted by
\begin{equation}
   \text{vec}(\bar{M}^*_{c k}) = T^t \cdot \text{vec}(\bar{M}_{c k} \odot (1 - \mathcal{B})),
   \label{eq:rw_prop}
\end{equation}
where $t$ denotes the number of iterations, $\odot$ is the Hadamard product, and vec$(\cdot)$ means vectorization.
We penalize scores of boundary pixels by multiplying $(1 - \mathcal{B})$ since those isolated pixels do not propagate their scores to neighbors and have overly high scores compared to the others in consequence.
Then an instance segmentation label is generated by choosing the combination of $c$ and $k$ that maximizes $\bar{M}_{ck}^*(\mathbf{x})$ 
for each pixel $\mathbf{x}$. 
If the maximum score is less than bottom 25\%, the pixel is regarded as background.

\section{Experiments}
\label{sec:experiments}

The effectiveness of our framework is demonstrated on the PASCAL VOC 2012 dataset~\cite{Everingham}, where our framework generates pseudo labels for training images and trains a fully supervised model with the images and their pseudo labels.
We evaluate the quality of our pseudo labels as well as the performance of the model trained with them.
The evaluation is done for both instance segmentation and semantic segmentation since our pseudo labels can be used to train semantic segmentation models as well.

\subsection{Experimental Setting}

\noindent \textbf{Dataset:} 
We train and evaluate our framework on the PASCAL VOC 2012~\cite{Pascalvoc} dataset. 
Although the dataset contains labels for semantic segmentation and instance segmentation, we only exploit image-level class labels. 
Following the common practice, the training set is expanded by adding image set proposed in~\cite{Hariharan}. 
In total, 10,582 images are used for training, and 1,449 images are kept for validation.

\noindent \textbf{Hyperparameter Settings:}
The radius that limits the search space of pairs $\gamma$ in \Eq{pair_set} is set to 10 when training, and reduced to 5 at inference for conservative propagation. 
The number of random walk iterations $t$ in \Eq{rw_prop} is fixed to 256. 
The hyperparameter $\beta$ in \Eq{trans_mat} is set to 10. 
The iterative update of $\mathcal{D}$ in \Eq{displacement_update} is done 100 times.

\noindent \textbf{Network Parameter Optimization:}
We adopt the stochastic gradient descent for network optimization.
Learning rate is initially set to 0.1, and decreases at every iteration with polynomial decay~\cite{Liu2015ParseNetLW}.
The backbone of IRNet is frozen during training, and gradients that displacement field branch receives are amplified by a factor of 10. 

\noindent \textbf{Comparison to AffinityNet:}
For a fair comparison, we modified AffinityNet~\cite{affinitynet} by replacing its backbone with ResNet50 as in our IRNet. 
Then we compare IRNet with the modified AffinityNet in terms of the accuracy of pseudo segmentation labels (Table~\ref{tab:synth_segm_anno_acc}) and performance of DeepLab~\cite{deeplab_v2} trained with these pseudo labels (Table~\ref{tab:comparison_segm}).

\begin{table}[!t] \small
\centering
\begin{tabular}{l|c}
    \hline
    Method & mIoU \\
    \hline
    CAM & 8.6 \\
    CAM + Class Boundary & 34.1 \\
    CAM + Displacement Field + Class Boundary (Ours) & 37.7 \\
    \hline
\end{tabular}
\vspace{0.1mm}
\caption{Quality of our pseudo instance segmentation labels in $\text{AP}^r_\text{50}$, evaluated on the PASCAL VOC 2012 \emph{train} set.}
\label{tab:synth_inst_anno_acc}
\end{table}

\begin{table}[!t] \small
\centering
\begin{tabular}{c||c|c}
    \hline
    CAM & Prop. w/ AffinityNet~\cite{affinitynet} & Prop. w/ IRNet (Ours) \\
    \hline
    48.3 & 59.3 & 66.5 \\
    \hline
\end{tabular}
\vspace{0.1cm}
\caption{Quality of pseudo semantic segmentation labels in mIoU, evaluated on the PASCAL VOC 2012 \emph{train} set. 
``Prop'' means the semantic propagation using predicted affinities. 
}
\vspace{-3mm}
\label{tab:synth_segm_anno_acc}
\end{table}

\subsection{Analysis of Pseudo Labels}

\noindent \textbf{Instance Segmentation labels:}
A few qualitative examples of pseudo instance segmentation labels are presented in \Fig{qualitative_labels}, and the contribution of each branch of IRNet to the quality of the labels is analyzed in \Tbl{synth_inst_anno_acc}.
In the case of ``CAM'' in \Tbl{synth_inst_anno_acc}, we directly utilize raw CAMs to generate pseudo labels by thresholding their scores and applying connected component analysis while assuming that there are no instances of the same class attached to each other.
In the case of ``CAM + Class Boundary'' in \Tbl{synth_inst_anno_acc}, pseudo labels are obtained in the same manner, but we enhance CAMs by the semantic propagation based on the class boundary map before generating pseudo labels.
We evaluated the performance of each method in terms of average precision (AP).
For evaluating APs, the score of each detected instance is given as the maximum class score within its mask.
As shown in the table, exploiting a class boundary map effectively improves the quality of pseudo labels by more than 25\% as it helps to recover the entire areas of objects missing in CAMs.
Exploiting a displacement field further improves the performance by 3.6\% as it helps to distinguish different instances of the same class.

\noindent \textbf{Semantic Segmentation Labels:}
A reduced version of our framework, which skips the instance-wise CAM generation step, produces pseudo labels for semantic segmentation.
In this aspect, we compare our framework with the previous state-of-the-art in semantic segmentation label synthesis, AffinityNet~\cite{affinitynet}, in terms of mean Intersection-over-Union (mIoU).
Similar to ours, AffinityNet also conducts the semantic propagation to enhance CAMs using predicted pairwise semantic affinities. 
\Tbl{synth_segm_anno_acc} compares the quality of our pseudo segmentation labels to that of AffinityNet~\cite{affinitynet}.
The accuracy of our pseudo labels is substantially higher than that of AffinityNet thanks to the superior quality of pairwise semantic affinities predicted by IRNet.

\begin{table}[!t] \footnotesize
\centering
\begin{tabular}{lcc|cc}
\hline
Method & Sup. & Extra data / Information & $\textrm{AP}^r_\textrm{50}$ & $\textrm{AP}^r_\textrm{70}$ \\
\hline
PRM~\cite{PRM} \raggedright                  & $\mathcal{I}$ & MCG~\cite{MCG} & 26.8 & - \\
SDI~\cite{SDI} \raggedright                  & $\mathcal{B}$ & BSDS~\cite{BSDS} & 44.8 & - \\
\hline
SDS~\cite{Sds} \raggedright & $\mathcal{F}$ & MCG~\cite{MCG} & 43.8 & 21.3 \\
MRCNN~\cite{mask_rcnn} \raggedright & $\mathcal{F}$ & MS-COCO~\cite{Mscoco} & 69.0 & - \\ 
\hline
\bf{Ours-ResNet50} \raggedright & $\mathcal{I}$ & - & 46.7 & 23.5 \\
\hline
\end{tabular}
\vspace{0.1mm}
\caption{Instance segmentation performance on the PASCAL VOC 2012 \emph{val} set. The supervision types (Sup.) indicate: $\mathcal{I}$--image-level label, $\mathcal{B}$--bounding box, and $\mathcal{F}$--segmentation label.}
\label{tab:comparison_inst}
\end{table}



\begin{table}[!t] \footnotesize
\centering
\begin{tabular}{lcc|cc}
\hline
Method & Sup. & Extra Data / Information & \emph{val} & \emph{test} \\
\hline
SEC~\cite{sec} \raggedright & $\mathcal{I}$ & - & 50.7 & 51.7 \\
AffinityNet~\cite{affinitynet} \raggedright      & $\mathcal{I}$ & -     & 58.7 & - \\
\hline
PRM~\cite{PRM} \raggedright                  & $\mathcal{I}$ & MCG~\cite{MCG} & 53.4 & - \\ 
CrawlSeg~\cite{Hong2017_webly} \raggedright      & $\mathcal{I}$ & YouTube Videos     & 58.1 & 58.7 \\
MDC~\cite{Wei_2018_CVPR} \raggedright & $\mathcal{I}$ & Ground-truth Backgrounds     & 60.4 & 60.8 \\
DSRG~\cite{Huang_2018_CVPR} \raggedright & $\mathcal{I}$ & MSRA-B~\cite{LiuCVPR07} & 61.4 & 63.2 \\
\hline
ScribbleSup~\cite{scribblesup} \raggedright  & $\mathcal{S}$ & - & 63.1 & -    \\
BoxSup~\cite{Boxsup} \raggedright            & $\mathcal{B}$ & - & 62.0 & 64.6 \\
SDI~\cite{SDI} \raggedright        & $\mathcal{B}$ & BSDS~\cite{BSDS} & 65.7 & 67.5 \\
\hline
Upperbound \raggedright & $\mathcal{F}$ & - & 72.3 & 72.5 \\
\bf{Ours-ResNet50} \raggedright                       & $\mathcal{I}$ & - & 63.5 & 64.8 \\
\hline
\end{tabular}
\vspace{0.1mm}
\caption{Semantic segmentation performance on the PASCAL VOC 2012 \emph{val} and \emph{test} sets. The supervision type (Sup.) indicates: $\mathcal{I}$--image-level label, $\mathcal{B}$--bounding box, $\mathcal{S}$--scribble, and $\mathcal{F}$--segmentation label.}
\label{tab:comparison_segm}
\vspace{-1mm}
\end{table}

%
\begin{figure*}[!ht]
\begin{center}
\includegraphics[width=1.00 \linewidth] {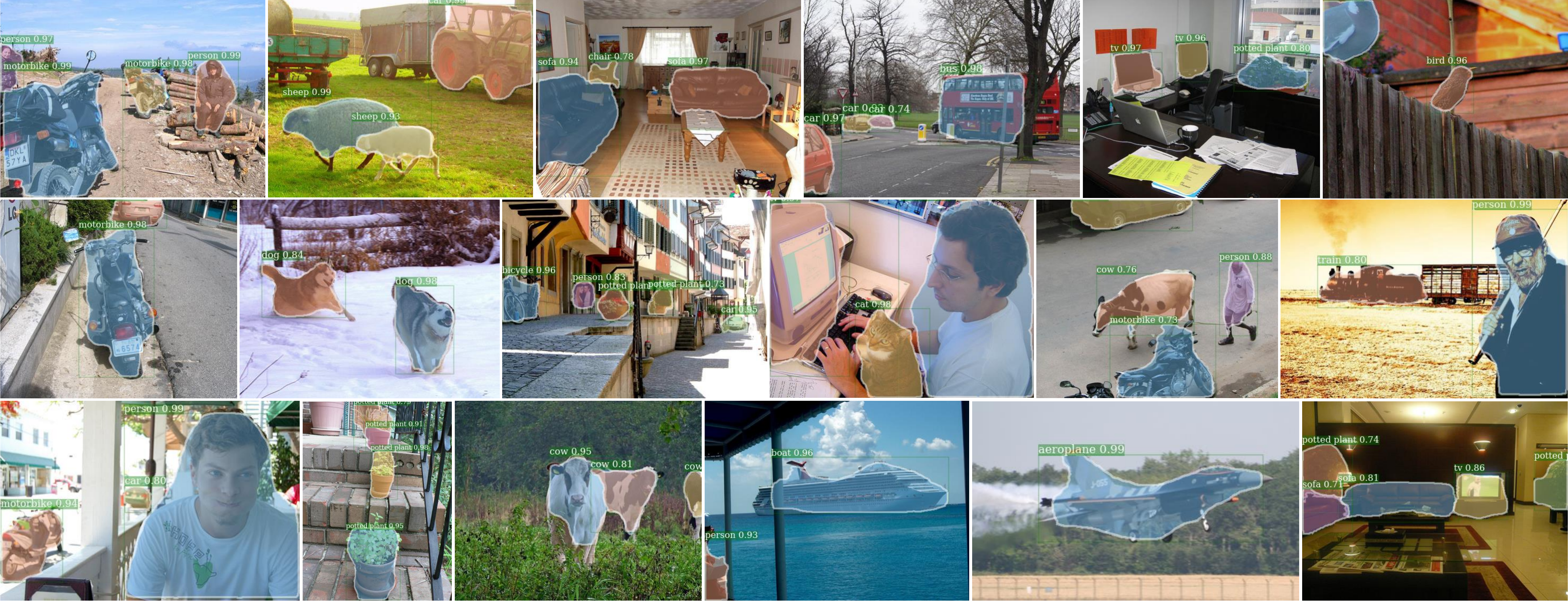}
\end{center}
\vspace{-0.4cm}
\caption{Qualitative results of our instance segmentation model on the PASCAL VOC 2012 \emph{val} set.
} 
\label{fig:qualitative_inst}
\end{figure*}
%
\begin{figure*}[!ht]
\begin{center}
\includegraphics[width=1.00 \linewidth] {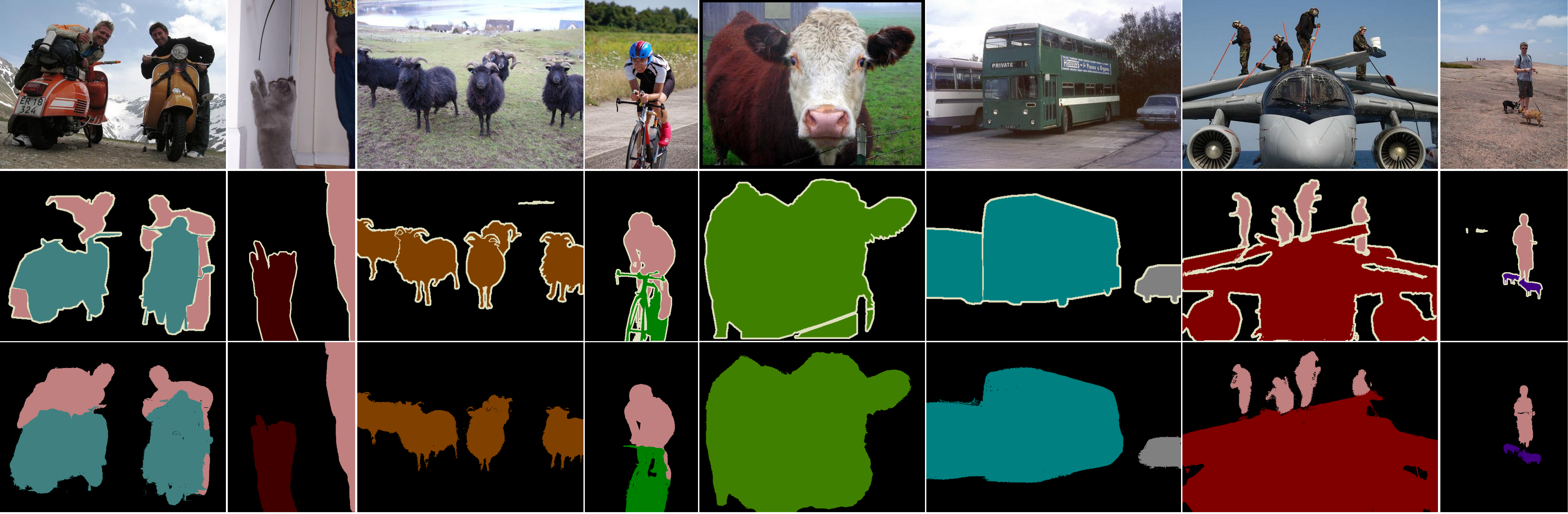}
\end{center}
\vspace{-0.4cm}
\caption{Qualitative results of smenatic segmentation on the PASCAL VOC 2012 \emph{val} set. (\emph{top}) Input images. (\emph{middle}) Groundtruth semantic segmentaton. (\emph{bottom}) Results of Ours-ResNet50.
} 
\label{fig:qualitative_segm}
\end{figure*}
%
\subsection{Mask R-CNN for Instance Segmentation}
We evaluate the performance of an instance segmentation network trained with pseudo labels generated by our framework.
For evaluation, we adopt Mask R-CNN~\cite{mask_rcnn}, which is one of the state-of-the-art instance segmentation networks, with ResNet-50-FPN~\cite{lin2017_fpn} as its backbone.
\Fig{qualitative_inst} shows qualitative results of the Mask-RCNN trained with our pseudo labels, and \Tbl{comparison_inst} compares its performance to those of previous approaches in $\text{AP}^r$\footnote{$\text{AP}^r$ means average precision of masks at different IoU thresholds.}~\cite{Sds}.
As shown in \Tbl{comparison_inst}, ours largely outperforms PRM~\cite{PRM}, which is the state-of-the-art that also uses image-level supervision.
Our approach even outperforms SDI~\cite{SDI}, which uses bounding box supervision, by 1.9\%, and SDS~\cite{Sds}, which uses full supervision, by 2.9\% in $\text{AP}^r_{\text{50}}$.



\subsection{DeepLab for Semantic Segmentation}
We further explore the effectiveness of our framework by training DeepLab v2-ResNet50~\cite{deeplab_v2} with our pseudo semantic segmentation labels.
\Fig{qualitative_segm} visualizes semantic segmentation results obtained by our approach and \Tbl{comparison_segm} compares ours with other weakly supervised approaches.
Our approach outperforms previous arts relying on the same level of supervision,
and is even competitive with BoxSup~\cite{Boxsup}, which utilizes stronger bounding box supervision.
Also it recovers 88\% of its fully supervised counterpart, the upper bound that it can achieve.


\section{Conclusion}
\label{sec:conclusion}

Weakly supervised instance segmentation with image-level supervision is a significantly ill-posed problem due to the lack of instance-specific information. 
To tackle this challenging problem, we propose IRNet, a novel CNN architecture that identifies individual instances and estimates their rough boundaries.
Thanks to the evidences provided by IRNet, simple class attentions can be significantly improved and used to train fully supervised instance segmentation models.
On the Pascal VOC 2012 dataset, models trained with our pseudo labels achieve the state-of-the-art performance in both instance and semantic segmentation.

\vspace{0.3cm}
{
\noindent \textbf{Acknowledgement:} This work was supported by Korea Creative Content Agency (KOCCA), Ministry of Culture, Sports, and Tourism (MCST) of Korea, Basic Science Research Program, and Next-Generation Information Computing Development Program through the National Research Foundation of Korea funded by the Ministry of Science, ICT (NRF-2018R1C1B6001223, NRF-2018R1A5A1060031, NRF-2017M3C4A7066316). It was also supported by the DGIST Start-up Fund Program (2018010071).
}

{\small
\bibliographystyle{ieee}
\bibliography{JW_weaksup_cvpr19}
}

\clearpage


\appendix
\section{Appendix}
\label{sec:appendix}

\newcolumntype{C}[1]{>{\centering\let\newline\\\arraybackslash\hspace{0pt}}p{#1}}

\def\ie{\emph{i.e.}}
\def\eg{\emph{e.g.}}
\def\etal{\emph{et al.}}
\def\wrt{\emph{w.r.t.}}

\definecolor{brown}{rgb}{0.65, 0.16, 0.16}
\definecolor{purp}{rgb}{0.65, 0.16, 0.65}






This appendix provides contents omitted in the regular sections for the sake of brevity.
\Sec{centroid} describes the centroid detection algorithm of \Sec{class_agnostic_instance_map} in more detail, 
and \Sec{segnet} introduces the instance and semantic segmentation models trained with our synthetic labels for the final evaluation.
Additional qualitative results are then presented in \Sec{results}.

\subsection{Details of the Centroid Detection Algorithm}
\label{sec:centroid}
As discussed in \Sec{class_agnostic_instance_map} of the main paper, a small group of neighboring pixels, instead of a single coordinate, are considered as a centroid in practice.
To this end, we first identify pixels whose displacement vectors in $\mathcal{D}$ have magnitudes smaller than a certain threshold, and consider them as candidate centroids. 
Specifically, the set of candidate centroids are defined as:
\begin{equation}
    \mathcal{C} = \big\{\mathbf{x} \mid \lVert \mathcal{D} (\mathbf x) \rVert_2 < 2.5 \big\} = \hat{\mathcal{C}}_1 \cup \hat{\mathcal{C}}_2 \cup \cdots \cup \hat{\mathcal{C}}_K,
\end{equation}
where $\hat{\mathcal{C}}_i$ is a connected component of pixels in $\mathcal{C}$ and $K$ is the number of connected components.
Then a class-agnostic instance map $I$ is obtained by assigning each pixel a connected component index in the following manner:
\begin{equation}
I(\mathbf{x}) = k, \ \ \textrm{if} \ \big(\mathbf{x} + \mathcal{D}(\mathbf{x}) \big) \in \hat{\mathcal{C}}_k, \ \ \forall \mathbf{x}.
\end{equation}

\subsection{Details of Our Segmentation Networks}
\label{sec:segnet}
As our framework aims to generate synthetic labels for instance and semantic segmentation, we evaluated the efficacy of our framework by learning fully supervised models for the two tasks with our synthetic labels. 
Specifically, we adopt Mask R-CNN~\cite{mask_rcnn} for instance segmentation and DeepLab v2~\cite{deeplab_v2} for semantic segmentation.
Both of them are first pretrained on ImageNet~\cite{Imagenet} then finetuned with the synthetic labels instead of groundtruth segmentation masks. 
The rest of this section describes details of the two models.


\subsubsection{Mask R-CNN for instance Segmentation}
We use Detectron~\cite{Detectron2018}, which is the official implementation of~\cite{mask_rcnn}, to implement Mask R-CNN~\cite{mask_rcnn} with ResNet-50-FPN~\cite{lin2017_fpn} as its backbone.
We directly adopt the default training setting given in the provided source code, except the number of training steps that is adjusted for better adaptation to the PASCAL VOC 2012 dataset~\cite{Pascalvoc}.


\subsubsection{DeepLab v2 for Semantic Segmentation}
We manually implement DeepLab v2~\cite{deeplab_v2} in PyTorch~\cite{pytorch}.
Its architecture consists of ResNet-50~\cite{resnet} followed by an atrous spatial pyramid pooling module~\cite{deeplab_v2}.
The training setting of ours is identical to that of the original model.
We also employ the ensemble of multi-scale prediction during evaluation. 
Specifically, a single input image is converted to a set of 8 images through resizing with 4 different scales \{0.5, 1.0, 1.5, 2.0\} and horizontal flip, and fed into the segmentation network so that the 8 outputs are aggregated by pixel-wise average pooling.

We also reproduce the performance of the fully supervised DeepLab v2, which is the \emph{upperbound} our segmentation model can achieve. 
Note that, as summarized in Table 4 of the main paper, \emph{upperbound} we measured is lower than the performance reported in the original paper~\cite{deeplab_v2} as we did not tune the parameters of dense CRF~\cite{Fullycrf} carefully. 
Thanks to the accurate segmentation labels synthesized in our framework, the DeepLab trained with our synthetic labels achieves 89.4\% of its fully supervised one on the PASCAL VOC 2012 \emph{test} set.



\subsection{More Qualitative Results of Our Approach}
\label{sec:results}
\begin{figure*}[!ht]
\hspace{0.7cm}Input Image\hspace{1.5cm} CAM \hspace{0.9cm} Displacement Field \hspace{0.4cm}Class Boundary\hspace{0.6cm} Instance Labels \hspace{0.6cm} Class Labels
\begin{center}
\includegraphics[width=0.98 \linewidth] {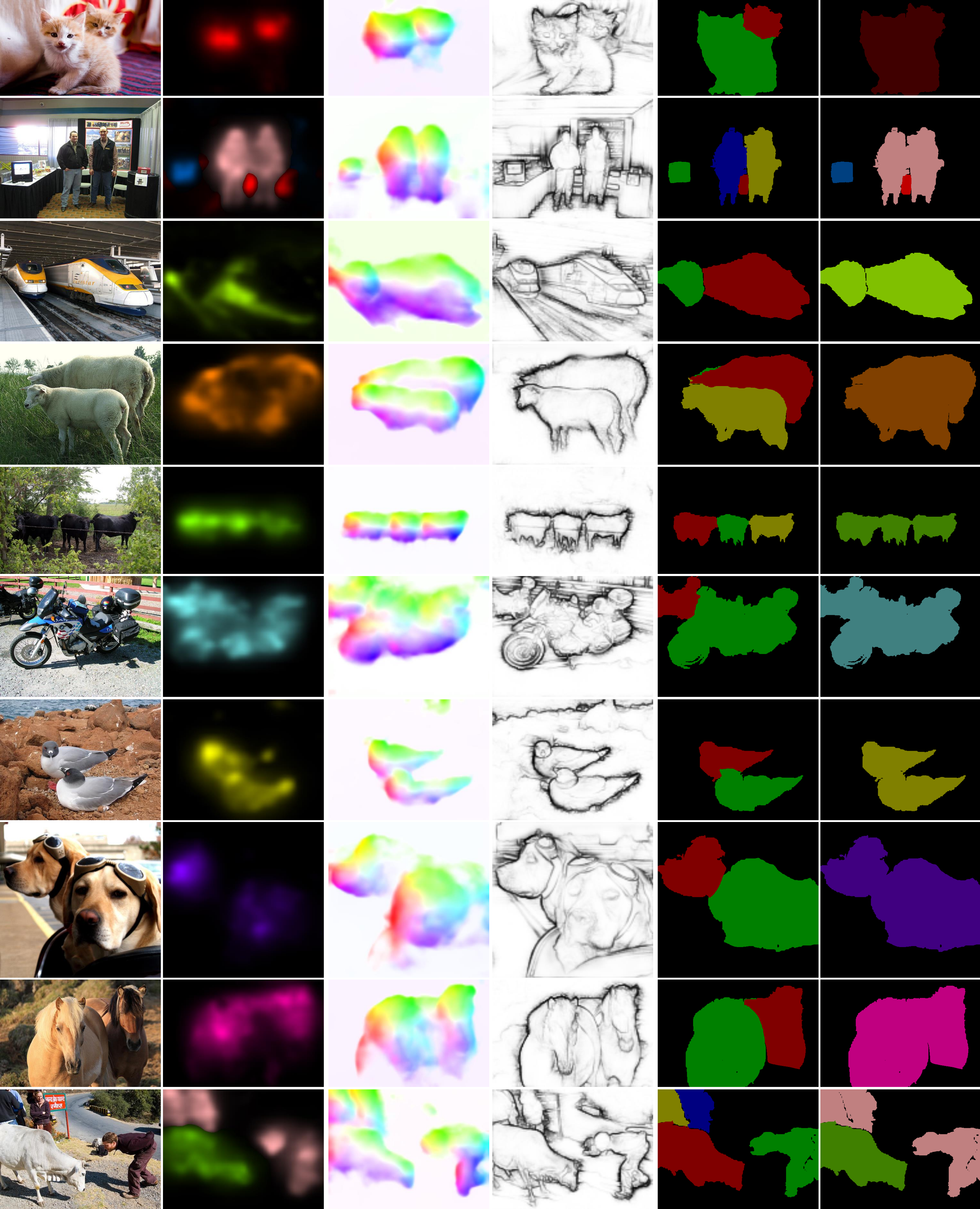}
\end{center}
\vspace{-0.2cm}
\caption{Qualitative results of our instance segmentation model on the PASCAL VOC 2012 \emph{train} set.
} 
\label{fig:qualitative_synth}
\end{figure*}

In this section, we provide additional qualitative results of our framework on the PASCAL VOC dataset. Although IRNet is trained with image-level supervision only, it successfully finds accurate class boundary and displacement field to instance centroids which are not directly available in CAMs, and synthesizes accurate instance segmentation masks from CAMs incorporating those two additional information as illustrated in \Fig{qualitative_synth}.

\Fig{qualitative_inst} and \Fig{qualitative_segm} show additional instance segmentation and semantic segmentation results of our models, respectively. Thanks to synthetic labels that are able to differentiate attached instances, our models not only find fine object shape, but also detect independent instances that are adjacent and of the same class.


%

%
%
\begin{figure*}[!ht]
\begin{center}
\includegraphics[width=0.97 \linewidth] {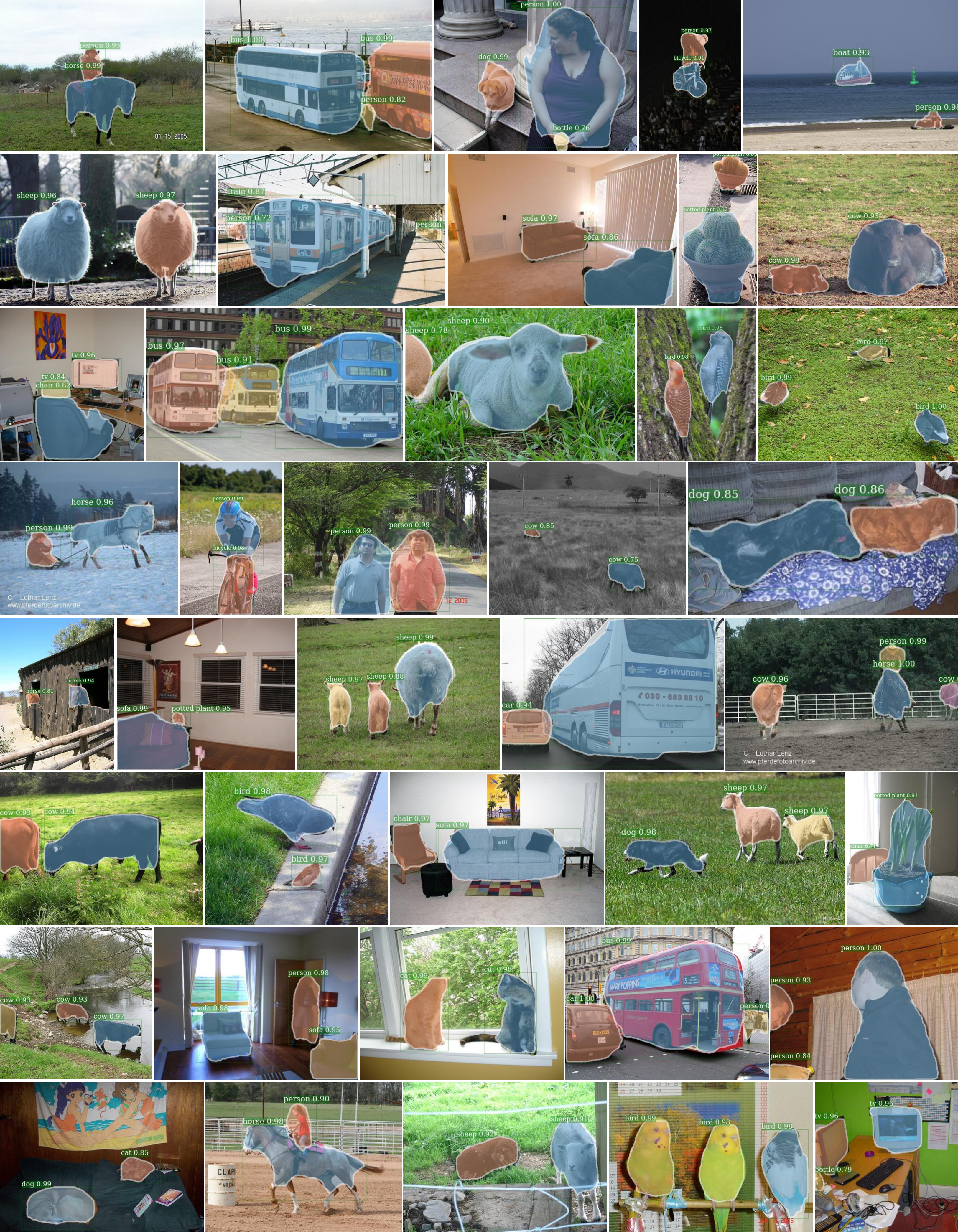}
\end{center}
\vspace{-0.2cm}
\caption{Qualitative results of our instance segmentation model on the PASCAL VOC 2012 \emph{val} set.
} 
\label{fig:qualitative_inst}
\end{figure*}
%

%
\begin{figure*}[!ht]
\hspace{0.5cm}Input Image\hspace{1.0cm} Ground-truth \hspace{1.5cm} Ours \hspace{1.7cm}Input Image\hspace{1.0cm} Ground-truth \hspace{1.5cm} Ours
\begin{center}
\includegraphics[width=1.00 \linewidth] {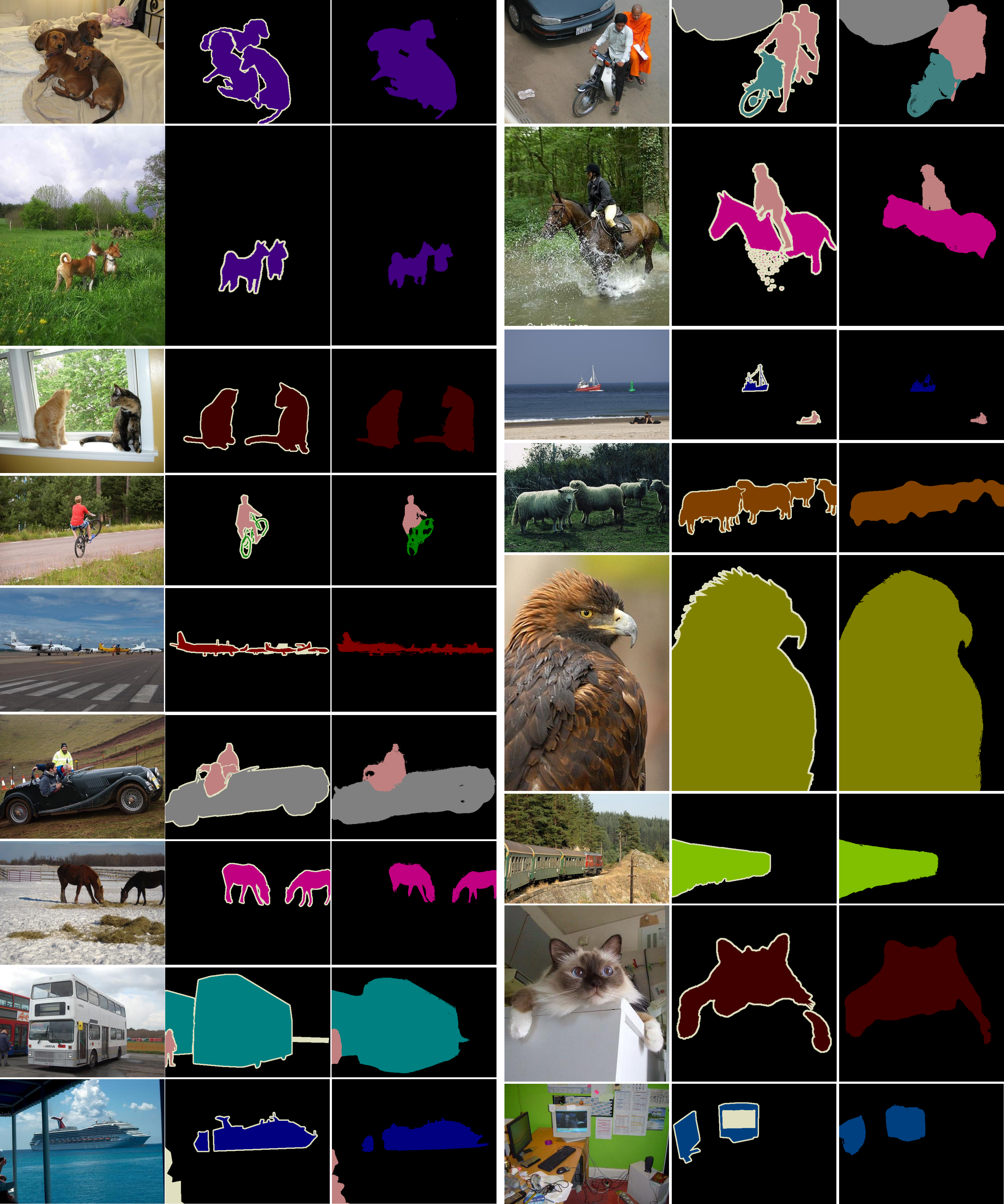}
\end{center}
\vspace{-0.2cm}
\caption{Qualitative results of our semantic segmentation model on the PASCAL VOC 2012 \emph{val} set.
} 
\label{fig:qualitative_segm}
\end{figure*}
%
$ $


\newpage



\end{document}